\documentclass[runningheads]{llncs}
\usepackage{graphicx}
\usepackage[misc]{ifsym}
\usepackage{tikz}
\usepackage{comment}
\usepackage{amsmath,amssymb} 
\usepackage{color}
\usepackage{multirow}
\usepackage{makecell}

\usepackage[breaklinks=true,
            bookmarks=false,
            colorlinks,
            linkcolor=red,       
            anchorcolor=blue,  
            citecolor=green, 
            ]{hyperref}

\usepackage[accsupp]{axessibility}  

\begin{document}
\pagestyle{headings}
\mainmatter
\def\ECCVSubNumber{968}  
\newcommand{\todo}[1]{{\color{red}{#1}}} 

\title{Metric Learning based Interactive Modulation for Real-World Super-Resolution}

\titlerunning{Metric Learning based Interactive Modulation for RWSR} 
\authorrunning{Chong Mou et al.}
\author{
Chong Mou$^{*1, 3}$, Yanze Wu$^{3}$, Xintao Wang$^\dagger$$^{3}$, Chao Dong$^{4, 5}$, Jian Zhang \Letter$^{ 1, 2}$, Ying Shan$^{3}$\\
$^1$Peking University Shenzhen Graduate School, Shenzhen, China\\
$^2$Peng Cheng Laboratory, Shenzhen, China\\
$^3$ARC Lab, Tencent PCG\\
$^4$Shenzhen Institutes of Advanced Technology, Chinese Academy of Sciences\\
$^5$Shanghai AI Laboratory\\
{\tt\small eechongm@gmail.com; $\{$yanzewu, xintaowang, yingsshan$\}$@tencent.com; chao.dong@siat.ac.cn; zhangjian.sz@pku.edu.cn}
}
\institute{ }
\maketitle
\let\thefootnote\relax\footnotetext{$^\dagger$ Project lead. \quad
\Letter \; Corresponding author.\\
$^*$ Chong Mou is an intern in ARC Lab, Tencent PCG.
}




\begin{abstract}
Interactive image restoration aims to restore images by adjusting several controlling coefficients, which determine the restoration strength. Existing methods are restricted in learning the controllable functions under the supervision of known degradation types and levels. They usually suffer from a severe performance drop when the real degradation is different from their assumptions. Such a limitation is due to the complexity of real-world degradations, which can not provide explicit supervision to the interactive modulation during training. However, how to realize the interactive modulation in real-world super-resolution has not yet been studied. In this work, we present a \textbf{M}etric Learning based Interactive \textbf{M}odulation for \textbf{Real}-World \textbf{S}uper-\textbf{R}esolution (\textbf{MM-RealSR}). Specifically, we propose an unsupervised degradation estimation strategy to estimate the degradation level in real-world scenarios. Instead of using known degradation levels as explicit supervision to the interactive mechanism, we propose a metric learning strategy to map the unquantifiable degradation levels in real-world scenarios to a metric space, which is trained in an unsupervised manner. Moreover, we introduce an anchor point strategy in the metric learning process to normalize the distribution of metric space. Extensive experiments demonstrate that the proposed MM-RealSR achieves excellent modulation and restoration performance in real-world super-resolution. 
Codes are available at \href{https://github.com/TencentARC/MM-RealSR}{https://github.com/TencentARC/MM-RealSR}.

\end{abstract}

\section{Introduction}
Image super-resolution (SR) is the task of recovering details of a high-resolution (HR) image from its low-resolution (LR) counterpart. Most SR methods \cite{srcnn,tsr2,tsr3,liu2021iterative,zhang2017collaborative} assume an ideal bicubic downsampling kernel, which is different from real-world degradations. This degradation mismatch makes those approaches unpractical in real-world scenarios. Recently, some attempts~\cite{sr1,bsrgan,realesrgan,sr3} are proposed to address real-world super-resolution (RWSR). Nevertheless, existing RWSR methods can only perform restoration with fixed one-to-one mapping. In other words, they lack the flexibility, \textit{i.e.}, interactive modulation~\cite{adafm,cresmd}, to alter the outputs by adjusting different restoration strength levels.

On the other hand, several modulation-based methods~\cite{adafm,cresmd,cresmd2,Coast} are proposed to enable controllable image restoration according to users' flavours. These modulation-based methods generally take the degradation level
as a part of the network inputs, and then construct mapping between the reconstruction result and the degradation level during training. 
During inference, users can simply adjust the value of the input degradation level, and the network will generate reconstruction results according to the corresponding restoration strength. 
However, existing controllable image restoration methods can only be trained on datasets with \textit{simple} degradation processes and \textit{known} degradation types/levels. 
%

\begin{figure*}[t]
\centering
\small 
\begin{minipage}[t]{.47\linewidth}
\centering
\includegraphics[width=1\columnwidth]{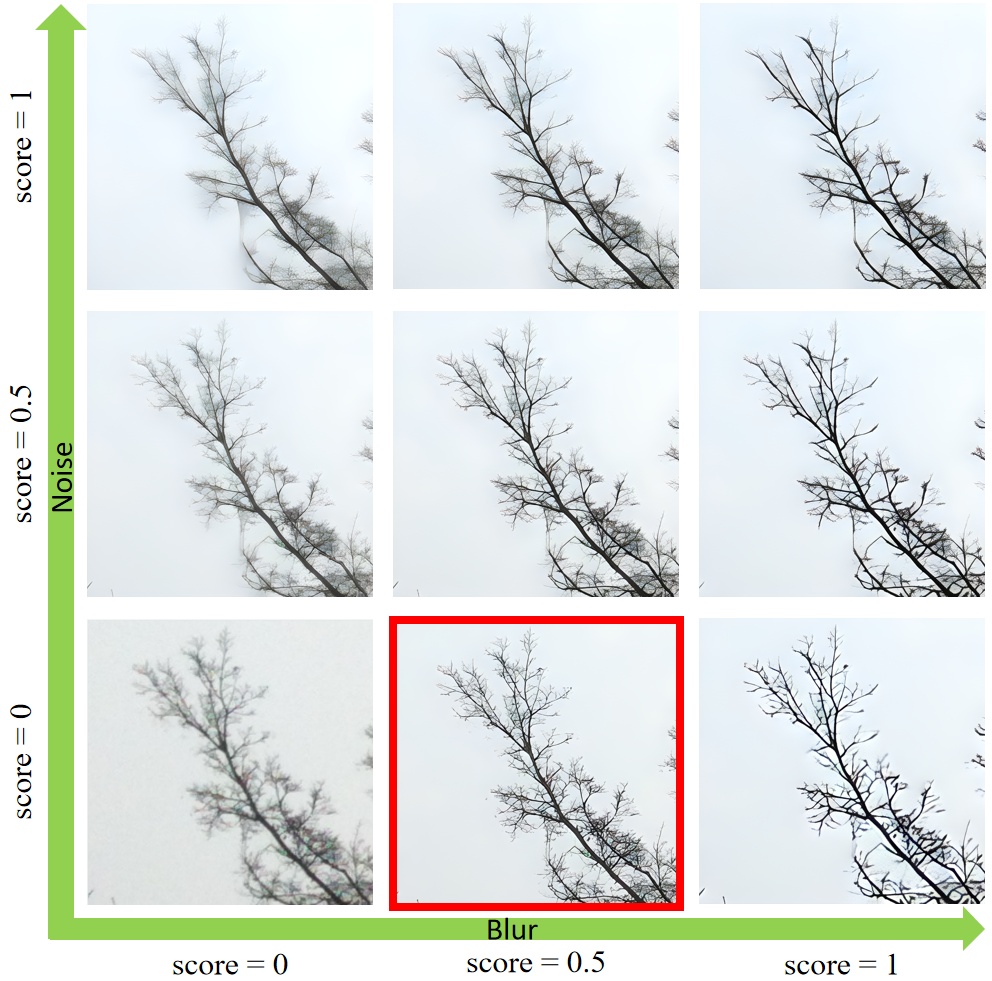}
\end{minipage}
\begin{minipage}[t]{.47\linewidth}
\centering
\includegraphics[width=1\columnwidth]{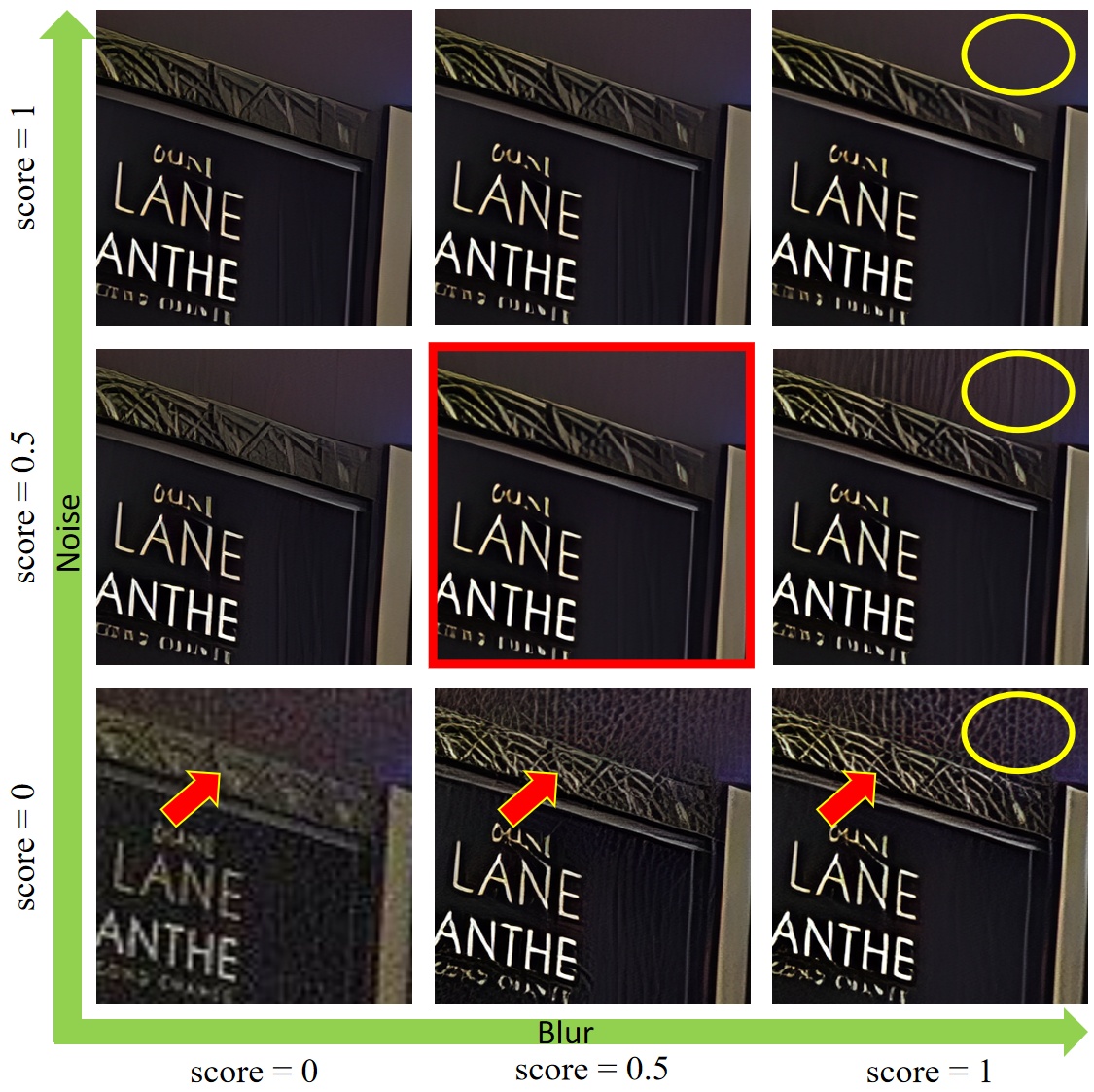}
\end{minipage}
\centering
\caption{Visual results of our interactive modulation method for real-world super-resolution ($\times 4$). Best results are labeled with red boxes. \textbf{(Zoom in for best view)}
}
\label{c_real_sy} 
\end{figure*}

In real-world scenarios, corrupted images usually contain mixture and complex degradations (\textit{e.g.}, blur~\cite{dgunet}, noise~\cite{zhang2012exploiting} and JPEG compression).
Such a complex real-world degradation process can be simulated by a random shuffle order~\cite{bsrgan} or a high-order degradation process~\cite{realesrgan}. Due to the mixtures of different degradation levels and complicated degradation types in high-order degradations, the  explicit  degradation  levels  can not reflect  the actual degradation effects in the corrupted images. For example, a corrupted image is degraded by a sequence of Gaussian blur with $\sigma_1$, Gaussian noise, and Gaussian blur with $\sigma_2$. We cannot know the final equivalent blur level on this degraded image.
%
This case becomes more complicated when the real-world degradation process involves more degradation types and more complex degradation combinations.
Therefore, existing modulation-based methods trained with known degradation types and levels can not get effective supervision in real-world settings.
How to extend the interactive modulation to RWSR is worth investigating and has not yet been studied. 

\begin{figure*}[t]
\centering
\small 
\begin{minipage}[t]{.85\linewidth}
\centering
\includegraphics[width=1\columnwidth]{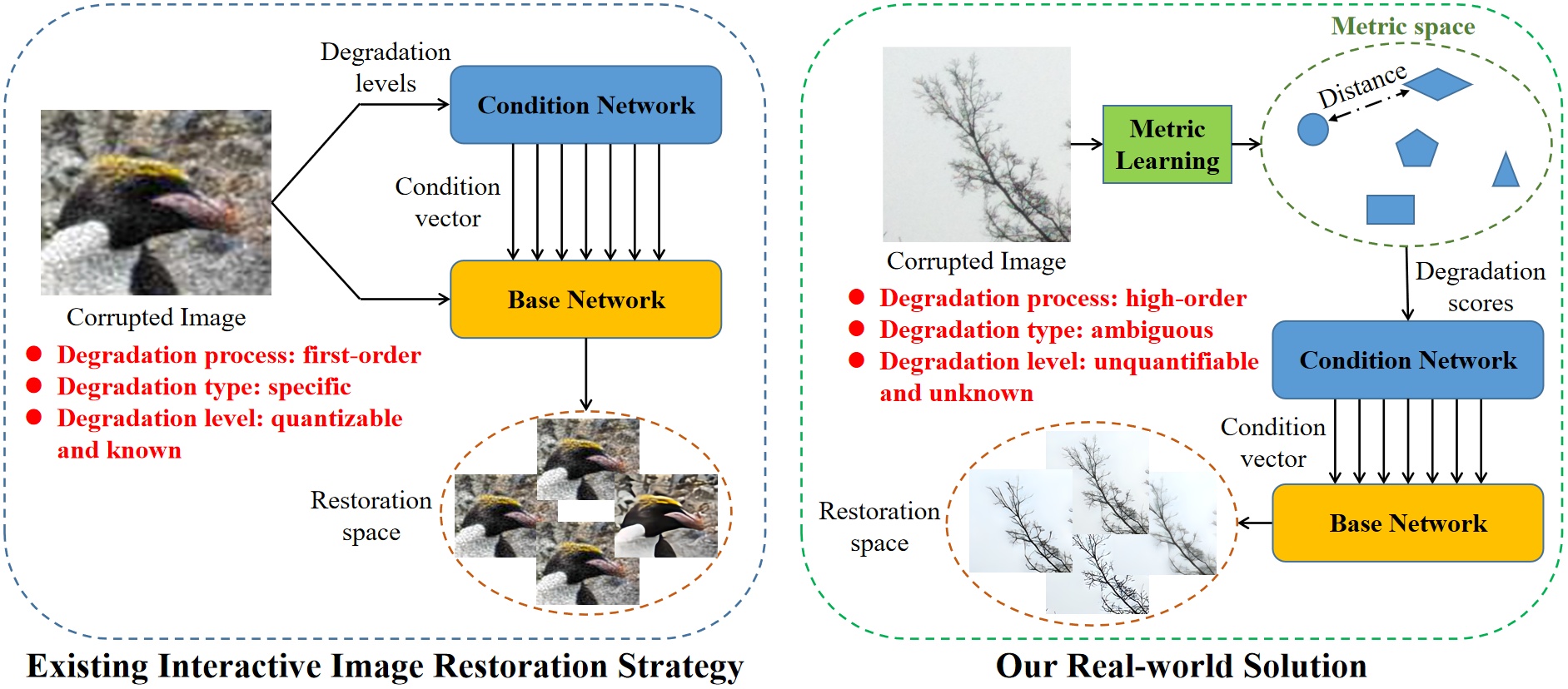}
\end{minipage}
\centering
\caption{Illustration of the existing interactive modulation strategy and our solution. Existing methods are confined to the sample with simple (\textit{e.g.}, the first-order) and known degradations. Our method is an unsupervised solution to solve the more challenging real-world problem.}
\label{fig_mcl} 
\end{figure*}

In this work, we present a \textbf{M}etric Learning based Interactive \textbf{M}odulation method for \textbf{Real}-World \textbf{S}uper-\textbf{R}esolution (MM-RealSR).
Specifically, 
we propose a metric learning scheme to map the unquantifiable degradation level to a ranking score (named \textit{degradation score} in this work) in a metric space. 
The degradation score generated from the metric space can reflect the relative strength of the degradation level and provide pseudo supervision to the interactive restoration mechanism.
A brief illustration and comparison to the existing interactive modulation strategy for image restoration are presented in Fig.~\ref{fig_mcl}. 
To restrict the learned degradation scores to a reasonable range and normalize the distribution of metric space, we further introduce an anchor point strategy in the metric learning process. Specifically, when the degradation score is zero, the network almost learns an identity mapping, while the network has the strongest restoration ability when the degradation score is one.
Equipped with such pseudo estimations for real-world degradations, our MM-RealSR can learn the controllable restoration by mapping the degradation scores to different restoration strengths.
Concretely, a condition network is used to generate controllable condition vectors to adjust the restoration quality. A base network is used to perform RWSR based on the input LR image and condition vectors. 
We summarize our contributions as follows. 
\textbf{1)} We propose a metric learning strategy to map unquantifiable degradation levels in real-world scenarios to a metric space in an  unsupervised manner. We further introduce an anchor point strategy
to normalize the distribution of metric space. 
\textbf{2)} Our proposed MM-RealSR is the first work to investigate the interactive modulation for RWSR.
\textbf{3)} Extensive experiments show that the proposed MM-RealSR achieves excellent modulation (\textit{e.g.}, Fig.~\ref{c_real_sy}) and restoration performance in RWSR.

\section{Related Work}

\subsection{Image Super-Resolution}
\label{SISR}
As a pioneer work, a three-convolutional-layer network is used in SRCNN~\cite{srcnn} to learn the LR-HR mapping for single image SR. Since then, motivated by the promising performance of CNN, several CNN-based SR methods have been proposed. 
Realizing the importance of the network depth, \cite{tsr4} designed a $20$-layer network with residual blocks. \cite{tsr7} then combined residual learning and dense connection to extend the network depth to $100$ layers. To enlarge the receptive field, \cite{nlsr1,nlsr2,dagl} proposed to utilize non-local self-similarity in SR. Recently, channel attention and second-order channel attention were further introduced by RCAN~\cite{rcan} and SAN~\cite{tsr6} to exploit feature correlation for improved performance. Some GAN-based methods~\cite{esrgan} were proposed to produce higher perceptual quality. Among existing methods, setting downsampling operation as a known prior is the most popular choice. However, they have poor generalization ability due to the domain gap between real-world LR images and clean-LR images used for training (typically clean bicubically downsampled images). 

There have been several attempts in blind SR \cite{liu2021blind,jiang2021towards} or real-world SR.
\cite{sr1,sr2} used a separate degradation estimation network to guide the blind SR process. Such methods usually consider simple synthetic degradations, and the accuracy of degradation representation largely affects the restoration quality. \cite{realsr,cam2} capture real-world training pairs through specific cameras followed by tedious alignments.
However, it is expensive to get such real-world training pairs, and it only works with specific imaging devices. \cite{sr3,sr4} directly learned from unpaired data with cycle consistency loss. Nevertheless, learning fine-grained degradations with unpaired data is challenging and usually unsatisfactory. \cite{bsrgan,realesrgan,smreal1,smreal2} generated training pairs as close to real data as possible, including various degradation factors and random/high-order degradation processes.

\subsection{Interactive Modulation for Image Restoration}
Most existing image restoration methods can only restore a specific image to a fixed result.
Controllable image restoration methods aim to solve this problem by allowing users to adjust the restoration strength. Some pioneer works (\textit{e.g.}, DNI~\cite{din} and AdaFM~\cite{adafm}) find that the learned filters of restoration models trained with different restoration levels are pretty similar at visual patterns. Based on this observation, DNI directly interpolated kernels to attain a smooth control of diverse imagery effects. AdaFM adopts a more efficient way by inserting AdaFM layers to change the statistics of filters. 
The work in CFSNet~\cite{cfsnet} adaptively learns the interpolation coefficients and uses them to couple intermediate features from the main branch and tuning branch. Recently, CResMD~\cite{cresmd} distinguished different degradation types in the modulation process and proposed a framework that accepts both corrupted images and their degradation information as input to realize multi-dimension modulation.
Following this strategy, \cite{cresmd2} proposed a GAN-based image restoration framework to produce higher perceptual quality with interactive modulation. However, existing modulation strategies can only be trained in a supervised manner to deal with simple and first-order degradations, which have limitations in real-world applications.  

\subsection{Metric Learning}
Metric learning~\cite{mcl} aims to measure the distance among samples while using an optimal distance metric for learning tasks. Instead of providing an explicit label, metric learning compares the distance among inputs to construct a metric space. In the last few years, deep learning and metric learning have been brought together to introduce the concept of deep metric learning~\cite{deepmcl}. Most of the existing deep metric learning methods can be roughly categorized based on loss functions: 1) contrastive loss~\cite{csl}; 2) triplet loss~\cite{tpl}; 3) margin ranking loss~\cite{mgl_2,mgl}. Contrastive loss and triplet loss have a similar formulation to close the distance between semantic-similar samples and enlarge the distance between dissimilar samples. Similar to the hinge loss in support vector machine (SVM)~\cite{svm}, margin ranking loss is designed for capturing the ranking relationship between input training samples.
In this paper, we utilize the margin ranking loss to construct the metric space of degradation levels, which drives our model to predict degradation levels in an unsupervised manner.

\begin{figure*}[t]
\centering
\small 
\begin{minipage}[t]{.8\linewidth}
\centering
\includegraphics[width=1\columnwidth]{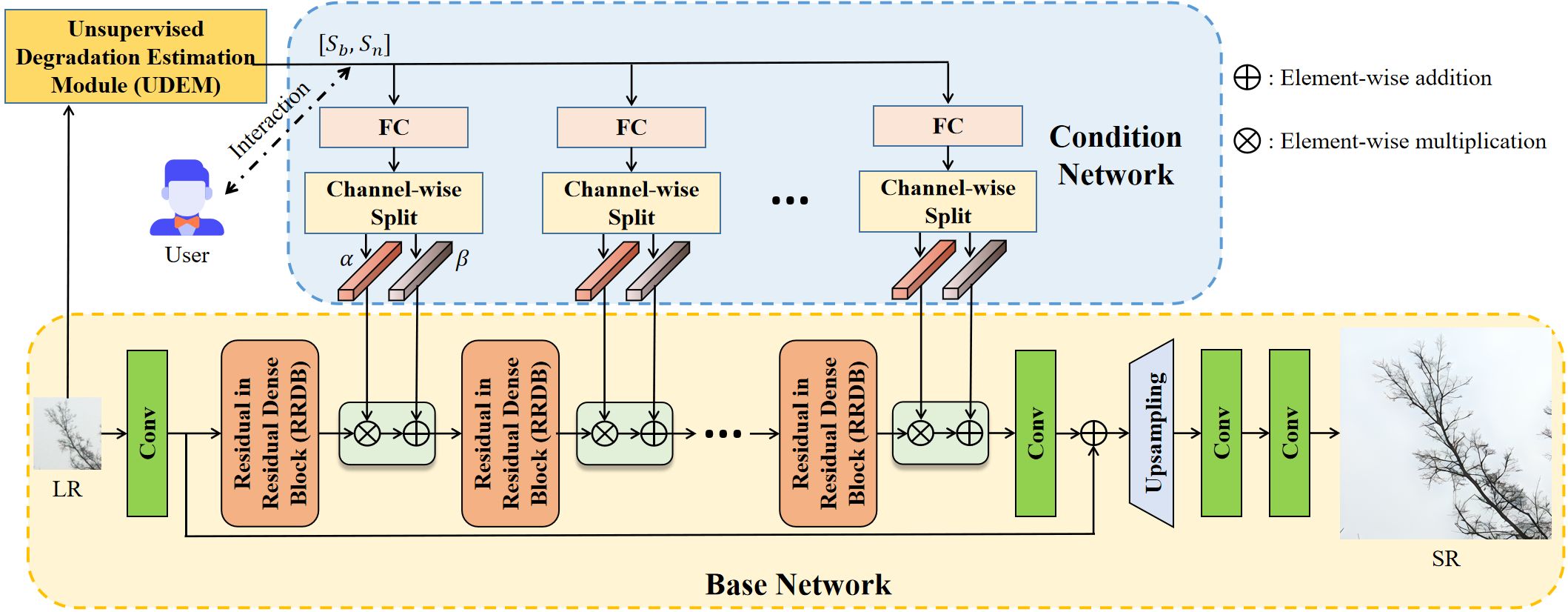}
\end{minipage}
\centering
\caption{Overview of our proposed metric learning based interactive modulation method for real-world super-resolution (MM-RealSR). It is composed of a base network, a condition network, and an unsupervised degradation estimation module.}
\label{overview} 
\end{figure*}

\section{Methodology}

\subsection{Overview}
The overview of our proposed method
(MM-RealSR) is presented in Fig.~\ref{overview}.
It is composed of a base network, a condition network, and an unsupervised degradation estimation module. The condition network takes the degradation scores as inputs to generate a condition vector. During inference, we can simply change the degradation scores of each degradation factor to adjust the restoration strength.
The base network takes condition vectors and degraded images as inputs to restore a clean result controlled by the condition vectors. The unsupervised degradation estimation module is used to map the unquantifiable degradation, which involve complex and real-world degradations, into a metric space. Each degraded image can find a degradation score in this metric space by the unsupervised degradation estimation module. Note that the degradation scores reflect the relative strength in the metric space instead of the absolute values of the degradation level. The estimated degradation scores can further provide pseudo supervision to train the condition network.  

\begin{figure*}[t]
\centering
\small 
\begin{minipage}[t]{.9\linewidth}
\centering
\includegraphics[width=1\columnwidth]{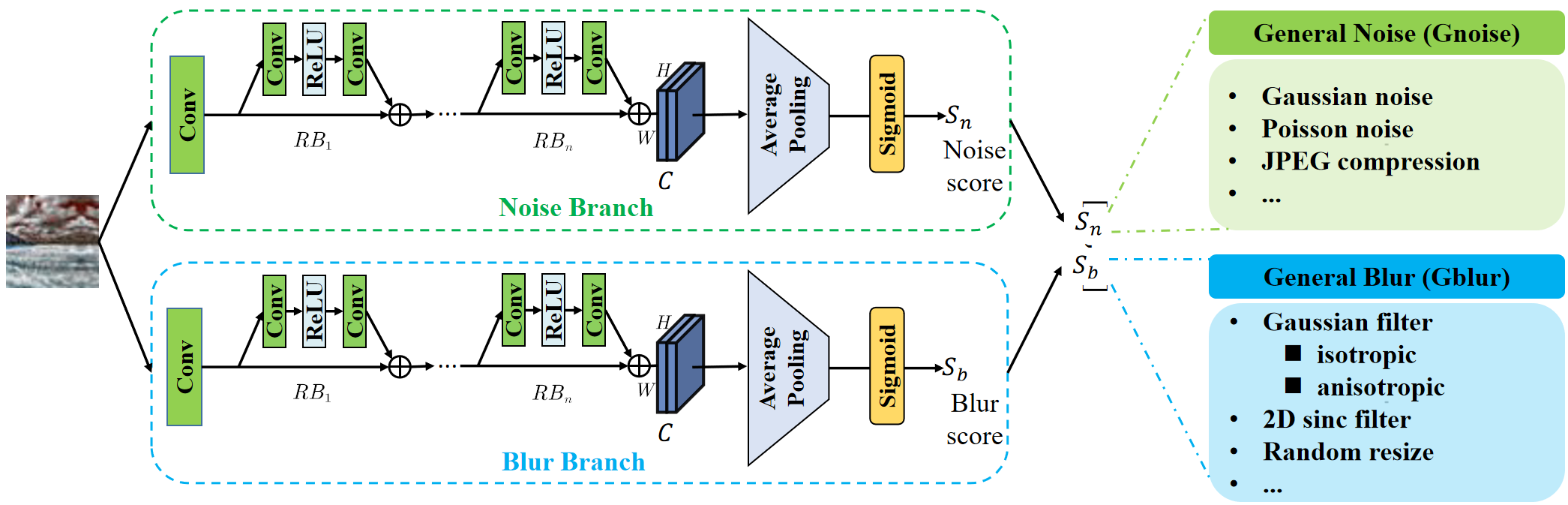}
\end{minipage}
\centering
\caption{The architecture of our proposed unsupervised degradation estimation module (UDEM). It is composed of two branches to estimate the degradation score of the general noise (Gnoise) and general blur (Gblur), respectively.}
\label{dem_arc} 
\end{figure*}

\begin{figure*}[t]
\centering
\small 
\begin{minipage}[t]{.9\linewidth}
\centering
\includegraphics[width=1\columnwidth]{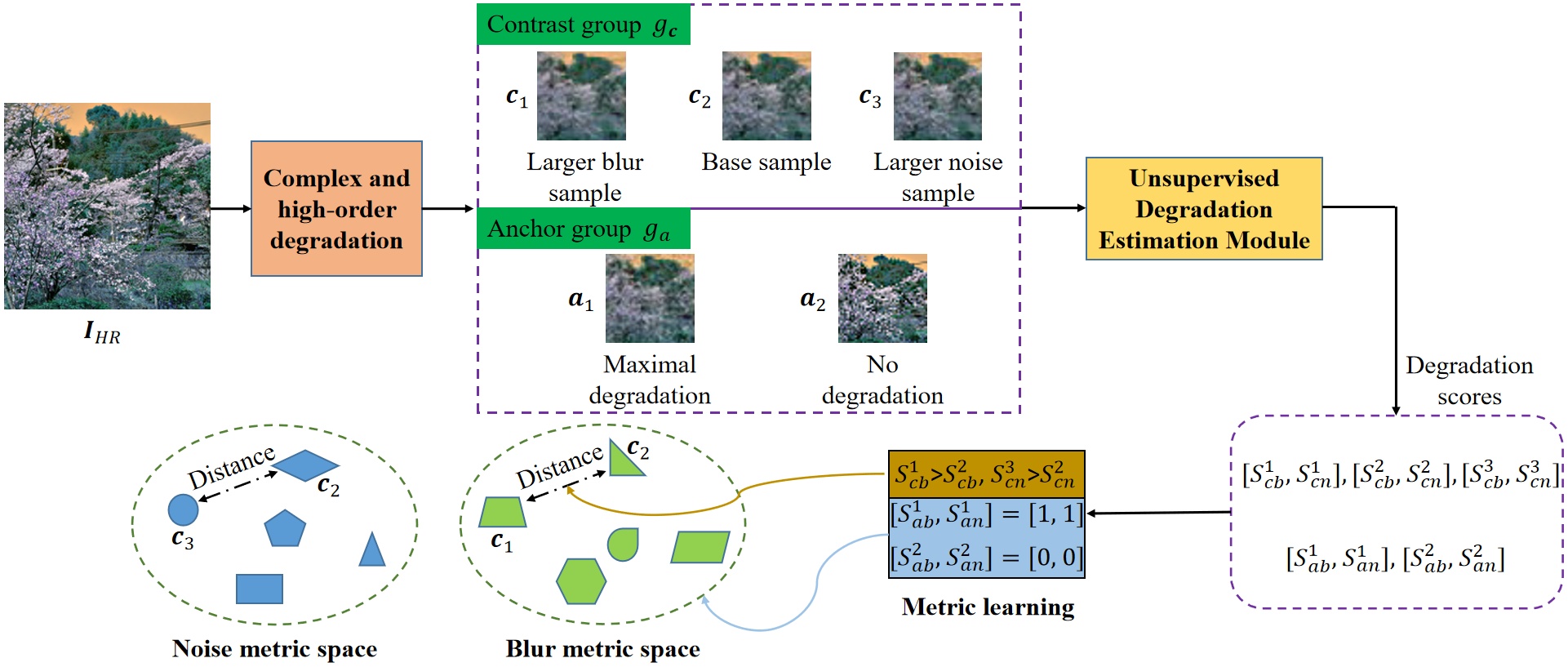}
\end{minipage}
\centering
\caption{The illustration of the optimization pipeline of our proposed unsupervised degradation estimation module. We utilize metric learning to construct a metric space for each degradation. We also apply two anchor points to restrict the learned degradation scores to a reasonable range and normalize the distribution of metric space.}
\label{dem} 
\end{figure*}

\subsection{Unsupervised Degradation Estimation Module (UDEM)}
\label{opt_mcl}
\textbf{Generalized degradation factors for controllable dimensions.}
Before we delve into the details of the unsupervised degradation estimation module (UDEM), we first present the controllable dimensions for real-world interactive modulation.
Recall that we want to investigate the interactive modulation under real-world degradations. 
We adopt the high-order degradation process~\cite{realesrgan,bsrgan} to simulate real-world degradations. 
In this setting, there are different degradation types (\textit{e.g.}, Gaussian noise, Poisson noise, anisotropic Gaussian blur) and thus we cannot simply determine the controllable dimensions according to their types.
Furthermore, due to the mixture of different degradation levels
in high-order degradations, the  explicit  degradation  levels  can not reflect  the actual degradation effects in the corrupted images. For example, a corrupted image is degraded by a sequence of Gaussian blur with $\sigma_1$, Gaussian noise, and Gaussian blur with $\sigma_2$. We cannot know the final equivalent blur levels on this degraded image.

Therefore, we need to re-define the controllable dimensions for real-world interactive modulation.
In this paper, we propose to adopt the \textit{general noise}  (Gnoise) and \textit{general blur} (Gblur) as the two controllable dimensions, as users usually focus on the visual effects brought by those two factors.
Specifically, general noise includes Gaussian noise, Poisson noise, and JPEG compression, as they all introduce unpleasant ``artifacts'' into images. 
Note that we classify JPEG compression artifacts as Gnoise, as the JPEG blocking artifacts are visually more like noise. 
The general blur includes Gaussian filter (isotropic/anisotropic), 2D sinc filter~\cite{realesrgan}, and random resize operation, as they all weaken image details.

\textbf{Module architecture.}
Our proposed UDEM is used to construct two metric spaces, corresponding to Gblur and Gnoise, respectively. Each metric space can convert the unquantifiable degradation level to a degradation score. A brief illustration of our UDEM is presented in Fig.~\ref{dem_arc}. Specifically, there are two prediction branches to predict the Gnoise score $S_{n}$ and Gblur score $S_{b}$, respectively. Each branch comprises several residual blocks end with an average pooling operation. 

\textbf{Metric learning pipeline.}
To construct the metric space with UDEM, we propose a metric learning pipeline, as illustrated in Fig.~\ref{dem}. 
Given the HR image $\mathbf{I}_{HR}$, we first corrupt it with complex and high-order degradations as~\cite{realesrgan} to simulate the real-world degraded images.
To train the metric space, we generate two groups of samples -- the contrast group $\mathbf{g}_{c}$ and anchor group $\mathbf{g}_{a}$. 
$\mathbf{g}_{c}$ is used to model the distance in the metric space by comparing the sample distance among this group, and $\mathbf{g}_{a}$ is to normalize the distribution of the metric space. 
Concretely, there are three samples in the contrast group $\mathbf{g}_{c}=\{ \mathbf{c}_1, \mathbf{c}_2, \mathbf{c}_3 \}$, in which $\mathbf{c}_1$ has larger Gblur degradation than $\mathbf{c}_2$, and $\mathbf{c}_3$ has larger Gnoise degradation than $\mathbf{c}_2$. There are two samples in the anchor group $\mathbf{g}_{a}=\{ \mathbf{a}_1, \mathbf{a}_2\}$. $\mathbf{a}_1$ has the maximal Gnoise degradation and Gblur degradation, and $\mathbf{a}_2$ does not have any Gnoise degradation and Gblur degradation. Correspondingly, there are two groups of degradation scores generated as:
\begin{align}
    \{ [S_{cn}^{1}, S_{cb}^{1}], [S_{cn}^{2}, S_{cb}^{2}], [S_{cn}^{3}, S_{cb}^{3}] \} &= \mathcal{F}_{UDEM}(\{ \mathbf{c}_1, \mathbf{c}_2, \mathbf{c}_3 \}) \\
    \{[S_{an}^{1}, S_{ab}^{1}], [S_{an}^{2}, S_{ab}^{2}]\} &= \mathcal{F}_{UDEM}(\{ \mathbf{a}_1, \mathbf{a}_2\}),
\end{align}
where $\mathcal{F}_{UDEM}(\cdot)$ represents the function of UDEM. 

We then adopt the margin ranking loss~\cite{mgl_2,mgl} to construct the Gnoise metric space and Gblur metric space.
The margin ranking loss is defined as:
\begin{align}
    \mathcal{L}_{ML} &= \frac{1}{2N}\sum_{i,j}^{N}\mathtt{max}(0,\gamma-\delta(\mathbf{s}_i, \mathbf{s}_j)\cdot (\hat{\mathbf{s}}_i - \hat{\mathbf{s}}_j)) \\
    \delta(\mathbf{s}_i, \mathbf{s}_j) &= \left\{
        \begin{aligned}
       1,\  \mathbf{s}_i \geq \mathbf{s}_j\\
       -1,\  \mathbf{s}_i < \mathbf{s}_j,
        \end{aligned}
        \right.
    \label{margin_loss}
\end{align}
where $\mathbf{s}_i$ and $\mathbf{s}_j$ are the ground truth scores. $\hat{\mathbf{s}}_i$ and $\hat{\mathbf{s}}_j$ are predicted scores. $\gamma$ is the margin parameter to constrain the distance between two samples. $N$ refers to the number of training samples. Note that during optimization, we do not need to know the explicit values of $\mathbf{s}_i$ and $\mathbf{s}_j$ but their relative values.
In our cases, two margin ranking losses are applied:
\begin{align}
    \mathcal{L}_{ML}^{n} &= \frac{1}{N}\sum_{i}^{N}\mathtt{max}(0,\gamma-(S_{cn}^{1\circledast i} - S_{cn}^{2\circledast i})) \\
    \mathcal{L}_{ML}^{b} &= \frac{1}{N}\sum_{i}^{N}\mathtt{max}(0,\gamma-(S_{cb}^{3\circledast i} - S_{cb}^{2\circledast i})),
\end{align}
where $\circledast i$ represents the $i$-$th$ training sample. 
Note that since we choose the larger Gblur and Gnoise samples for $\mathbf{c}_1$ and $\mathbf{c}_3$, we already know their relative values, thus omitting the $\delta(\cdot, \cdot)$ function.

In order to further restrict the learned degradation scores to a reasonable range and normalize the distribution of metric space, we introduce an anchor point strategy in the metric learning process. Specifically, when the degradation score is zero, the network is enforced to learns an identity mapping, while the network has the strongest restoration ability when the degradation score is one.
The introduced anchor loss ($\mathcal{L}_{AC}$) is defined as:
\begin{align}
&\mathcal{L}_{AC} = \mathcal{L}_{AC}^{u} + \mathcal{L}_{AC}^{l}\\
&\left\{
        \begin{aligned}
&\mathcal{L}_{AC}^{u} = \frac{1}{N}\sum_{i}^{N}(||S_{an}^{1\circledast i}-1||_2^2 + ||S_{ab}^{1\circledast i}-1||_2^2)\\
&\mathcal{L}_{AC}^{l} = \frac{1}{N}\sum_{i}^{N}(||S_{an}^{2\circledast i}||_2^2 + ||S_{ab}^{2\circledast i}||_2^2),
        \end{aligned}
        \right.
\label{loss_anchor}
\end{align}
where $||\cdot||_2^2$ represents the $\ell_2$ norm. $\mathcal{L}_{AC}^{u}$ and $\mathcal{L}_{AC}^{l}$ constraint the upper bound and lower bound of degradation scores.

\subsection{Base Network}
Our proposed MM-RealSR is a GAN-based solution for interactive RWSR. Thus, the base network consists of a generator \textit{G} and a discriminator \textit{D}. The generator \textit{G} accepts the input image and outputs the restored result, while the discriminator \textit{D} aims to discriminate the restored result and the ground truth.

We adopt the same generator as ESRGAN~\cite{esrgan} and Real-ESRGAN~\cite{realesrgan}. Specifically, it comprises several residual-in-residual dense blocks (RRDB), as shown in Fig.~\ref{overview}. At the end of the network, a pixel-shuffle upsampling and two convolution layers are applied for reconstruction. For discriminator, we apply the same architecture as  Real-ESRGAN, \textit{i.e.}, a U-Net design to better consider the realness values for each pixel.
Note that the restoration performance of Real-ESRGAN is the theoretical upper bound of our proposed MM-RealSR, as MM-RealSR needs to learn another objective for interactive modulation.

\subsection{Condition Network}
As shown in Fig.~\ref{overview}, the condition network is composed of several fully-connected (FC) layers, placed at the end of each RRDB. Each FC takes degradation scores $[S_n, S_b]$ as input and generates a condition vector $Z_i\in \mathbb{R}^{B\times 2C}$, where $B$ and $C$ represent the batch size and the number of channels. $i$ represents the modulation operation in the $i$-$th$ RRDB. Then, a channel-wise splitting operation divides $Z_i$ into two affine transformation parameters $\alpha_i,\ \beta_i \in \mathbb{R}^{B\times C}$. The modulation operation is performed through an affine transformation with $\alpha_i$ and $\beta_i$:
\begin{equation}
    \widetilde{\mathbf{F}}_i = \mathbf{F}_i\cdot \alpha_i + \beta_i,
\end{equation}
where $\mathbf{F}_i$ and $\widetilde{\mathbf{F}}_i$ refer to the output feature in the $i$-$th$ RRDB and the transformation results, respectively.

\subsection{Loss Function}
The proposed MM-RealSR constructs a conditional mapping in real-world super-resolution. The training objective is to drive our MM-RealSR to match this mapping relation, leading to two specific goals, \textit{i.e.}, interactive modulation mechanism and GAN-based image restoration. 
The interactive modulation mechanism is optimized with the metric learning strategy, as illustrated in Sec.~\ref{opt_mcl}, including two margin ranking losses ($\mathcal{L}_{ML}^{n}$, $\mathcal{L}_{ML}^{b}$) and two anchor losses ($\mathcal{L}_{AC}^{u}$, $\mathcal{L}_{AC}^{l}$).
The GAN-based image restoration is optimized with $\mathcal{L}_{1}$ loss, perceptual loss $\mathcal{L}_{per}$~\cite{l_p}, and GAN loss $\mathcal{L}_{GAN}$~\cite{l_g}. 
The total training objective is defined as:
\begin{equation}
    \mathcal{L} = \lambda_1 \mathcal{L}_{GAN} + \lambda_2 \mathcal{L}_{per} + \lambda_3 \mathcal{L}_{1} + \lambda_4(\mathcal{L}_{ML}^{n} + \mathcal{L}_{ML}^{b} + \mathcal{L}_{AC}^{u} + \mathcal{L}_{AC}^{l}),
    \label{loss}
\end{equation}
where $\{ \lambda_1,\ \lambda_2,\ \lambda_3,\ \lambda_4\}$ are hyper parameters to balance these loss items. In our implementation, we set these four weights as $\{ \lambda_1,\ \lambda_2,\ \lambda_3,\ \lambda_4\}=\{ 0.1,\ 1,\ 1,\ 0.05\}$.

\section{Experiment}
\subsection{Experimental Settings}
\subsubsection{Datasets.} Similar to ESRGAN~\cite{esrgan} and Real-ESRGAN~\cite{realesrgan}, we adopt DIV2K~\cite{div2k}, Flickr2K~\cite{flickr2k} and OutdoorSceneTraining~\cite{outdoorscene} datasets for training. For evaluation, we use the dataset provided in the challenge of Real-World Super-Resolution: AIM19~\cite{aim19}, and we also use a test set:
RealSRSet~\cite{realsr} modeling DLSR camera corruptions. In all cases, we use $\times 4$ upsampling.

\textbf{Training Details.}
Our training process consists of two stages. First, we pre-train the PSNR-oriented modulation model through Eq.~\ref{loss}, without perceptual loss $\mathcal{L}_{per}$ and GAN loss $\mathcal{L}_{GAN}$. In this stage, We train our system for $1000K$ iterations with the learning rate $2\times 10^{-4}$. The well-trained model is served as the starting point for the next GAN training. In the second stage, our system is optimized with Eq.~\ref{loss} for $400K$ iterations. The learning rate is fixed as $1\times 10^{-4}$. For optimization, we use Adam~\cite{adam} with $\beta_1=0.9$ and $\beta_2=0.99$. In both two stages, we set the batch size to $48$, with the input patch size being $64$.

\textbf{Evaluation Framework.}
The main contribution of this paper is extending the interactive modulation mechanism to RWSR with the help of metric learning. The evaluation includes three aspects. \textbf{First}, we evaluate the estimation ability of our UDEM to present that the proposed metric learning strategy can well perform degradation estimation in an unsupervised manner. \textbf{Second}, we evaluate the restoration performance without manually adjusting the degradation score estimated by UDEM. This aims to present that our MM-RealSR can achieve state-of-the-art RWSR performance. \textbf{Third}, we compare our MM-RealSR with the recent modulation SR methods to present that existing modulation methods can not deal with real-world scenarios.
Since AIM19 and RealSRSet datasets provide a paired validation set, we compute the learned perceptual image patch similarity (LPIPS)~\cite{lpips} and deep image structure and texture similarity (DISTS)~\cite{dists} as the evaluation items. We also adopt a non-reference image quality assessment (IQA) metric (\textit{i.e.,} NIQE~\cite{niqe}) for quantitative evaluation.

\subsection{Effectiveness of the Unsupervised Degradation Estimation}
\begin{figure*}[t]
\centering
\small 
\begin{minipage}[t]{.9\linewidth}
\centering
\includegraphics[width=1\columnwidth]{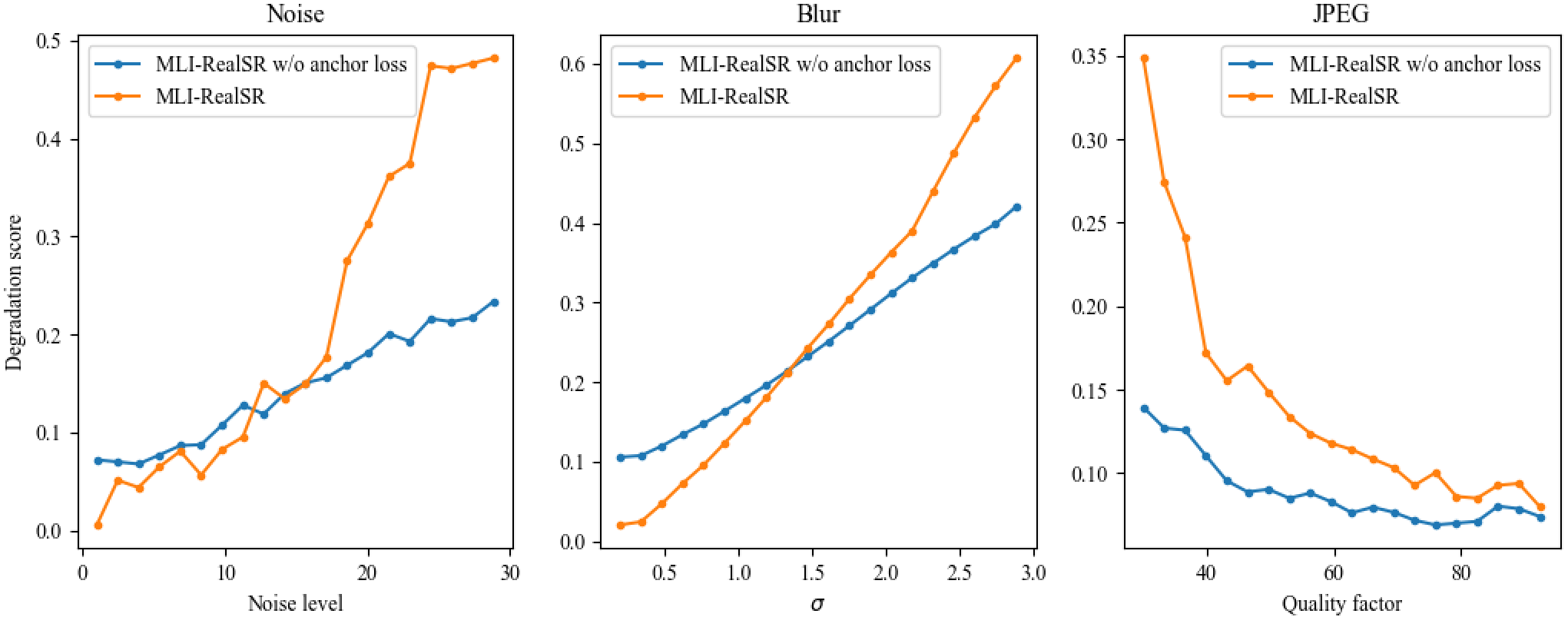}
\end{minipage}
\centering
\caption{The evaluation of our UDEM, including the model trained with and without the anchor loss. It shows that metric learning can generate good metric spaces to rank the degradation levels for various degradation types. The anchor loss can further normalize the space distribution and better distinguish degradation levels in a larger score range.}
\label{dist} 
\end{figure*}

As mentioned above, our degradation estimation module is optimized in an unsupervised manner to adapt the real-world degradation settings. To verify the effectiveness of this module, we first evaluate it on the synthesized data, corrupted by noise, blur, and JPEG compression degradations, respectively.
For the noise degradation, we choose the Gaussian noise with the noise level range $[1,\ 30]$. For the blur degradation, we choose the Gaussian blur, with the $\sigma$ range $[0.2, 3]$. For the JPEG compression, we set the quality factor ranging from $30$ to $95$.
We divide the degradation range of these three kinds of degradations equally into $20$ points and add them to natural images, respectively. The evaluation results are presented in Fig.~\ref{dist}, including the estimation performance of our proposed MM-RealSR and the same model trained without anchor loss (w/o anchor loss). One can see that with the increase of degradation level, the estimated degradation score increases monotonously. Note that for JPEG compression, the degradation artifact is weakened with the increase of the quality factor.

Additionally, the anchor loss allows the estimated score to have a larger dynamic range.
Therefore, our unsupervised degradation estimation module can accurately estimate the relative value from the corresponding metric space. The anchor loss (Eq.~\ref{loss_anchor}) can constrain the distribution of points in metric space.
In Fig.~\ref{dist}, there are some minor fluctuations for estimating noise and JPEG artifacts, while the curve is pretty smooth for the blur degradation.
We conjecture that estimating noise is more difficult than estimating blur, as distinguishing the textures on bricks, wood, and noisy artifacts are challenging.

\begin{figure*}[t]
\centering
\footnotesize
\begin{minipage}[t]{0.16\linewidth}
\centering
\includegraphics[width=1\columnwidth]{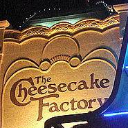}\\$S_b=0.0017$\\$S_n=0.4335$
\end{minipage}
\begin{minipage}[t]{0.16\linewidth}
\centering
\includegraphics[width=1\columnwidth]{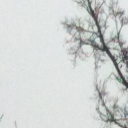}\\$S_b=0.5086$\\$S_n=0.0012$
\end{minipage}
\begin{minipage}[t]{0.16\linewidth}
\centering
\includegraphics[width=1\columnwidth]{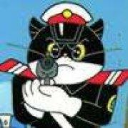}\\$S_b=0.3110$\\$S_n=0.3495$
\end{minipage}
\begin{minipage}[t]{0.16\linewidth}
\centering
\includegraphics[width=1\columnwidth]{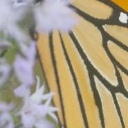}\\$S_b=0.3945$\\$S_n=0.1804$
\end{minipage}
\begin{minipage}[t]{0.16\linewidth}
\centering
\includegraphics[width=1\columnwidth]{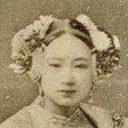}\\$S_b=0.3134$\\$S_n=0.7313$
\end{minipage}
\begin{minipage}[t]{0.16\linewidth}
\centering
\includegraphics[width=1\columnwidth]{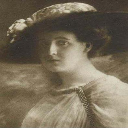}\\$S_b=0.7712$\\$S_n=0.3898$
\end{minipage}
\centering
\caption{Visualization of the degradation scoring ability of our unsupervised degradation estimation module (UDEM) on real-world data.
}
\label{v_score} 
\end{figure*}

Apart from the quantitative evaluation, we also visualize the scoring performance on the real-world data, as presented in Fig.~\ref{v_score}. We can find that the scoring results on the real-world data are reasonable and satisfactory. The first sample is noisy, and the second sample is blurry. We can observe that they obtain large scores on noise and blur, respectively, while they have a very low score on the other degradations. The fifth sample has large noise with some blur, and the predicted scores can also reflect such an observation. 
Note that the scoring for noise and blur is achieved in two separate metric spaces.
Therefore, the noise score and blur score are not suitable for strict corresponding, and there might be a slight difference with visual comparison.

\begin{figure*}[t]
\centering
\footnotesize
\begin{minipage}[t]{0.16\linewidth}
\centering
\includegraphics[width=1\columnwidth]{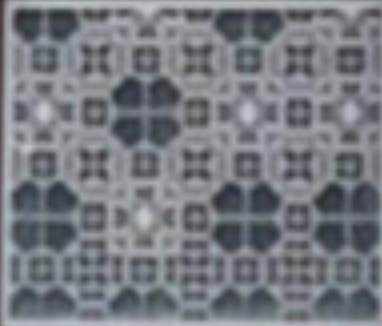}\\
\includegraphics[width=1\columnwidth]{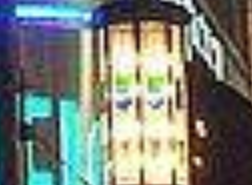}\\
\includegraphics[width=1\columnwidth]{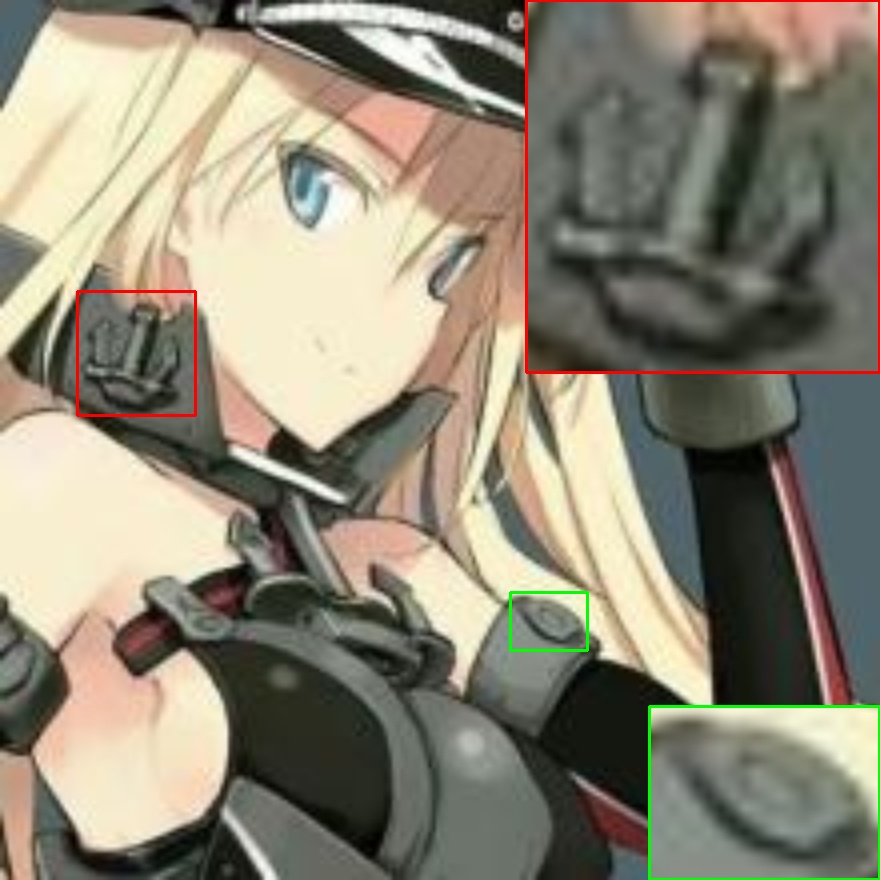}\\Bicubic $\times 4$
\end{minipage}
\begin{minipage}[t]{0.16\linewidth}
\centering
\includegraphics[width=1\columnwidth]{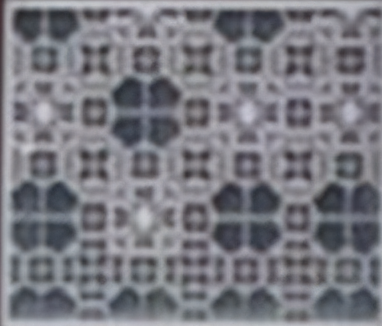}\\
\includegraphics[width=1\columnwidth]{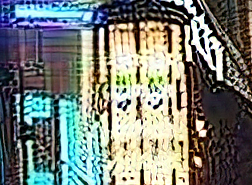}\\
\includegraphics[width=1\columnwidth]{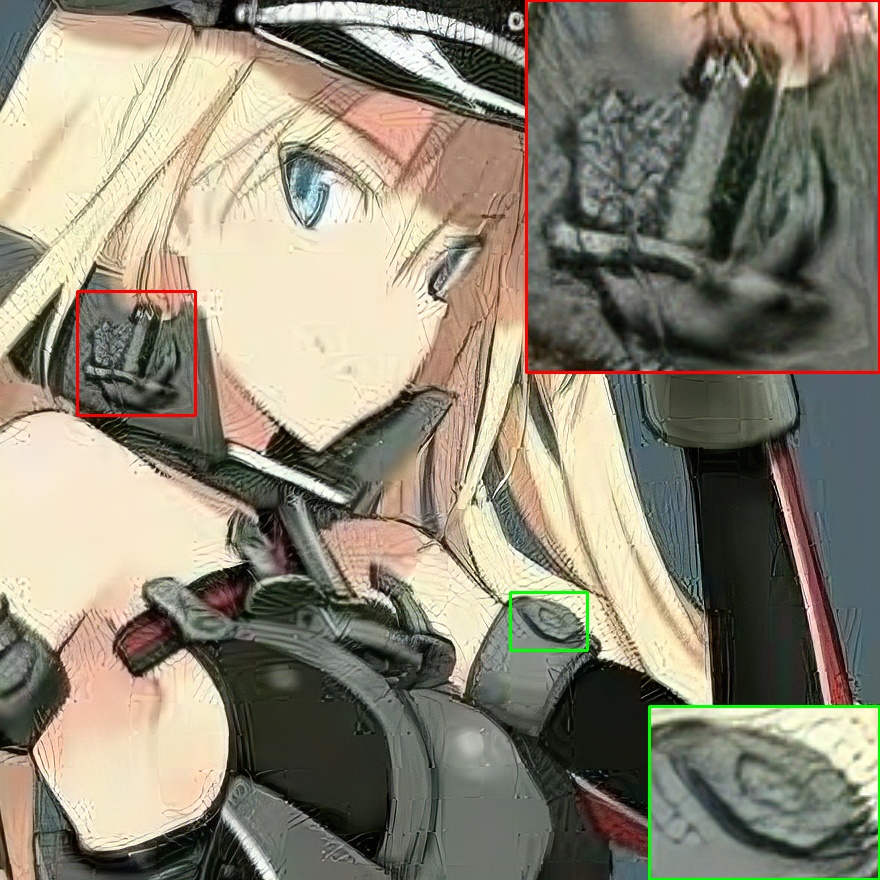}\\RealSR~\cite{smreal2}
\end{minipage}
\begin{minipage}[t]{0.16\linewidth}
\centering
\includegraphics[width=1\columnwidth]{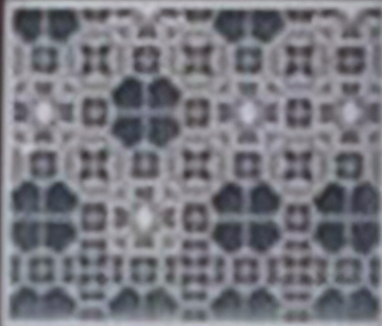}\\
\includegraphics[width=1\columnwidth]{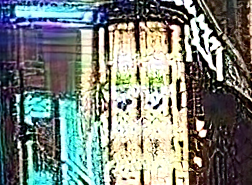}\\
\includegraphics[width=1\columnwidth]{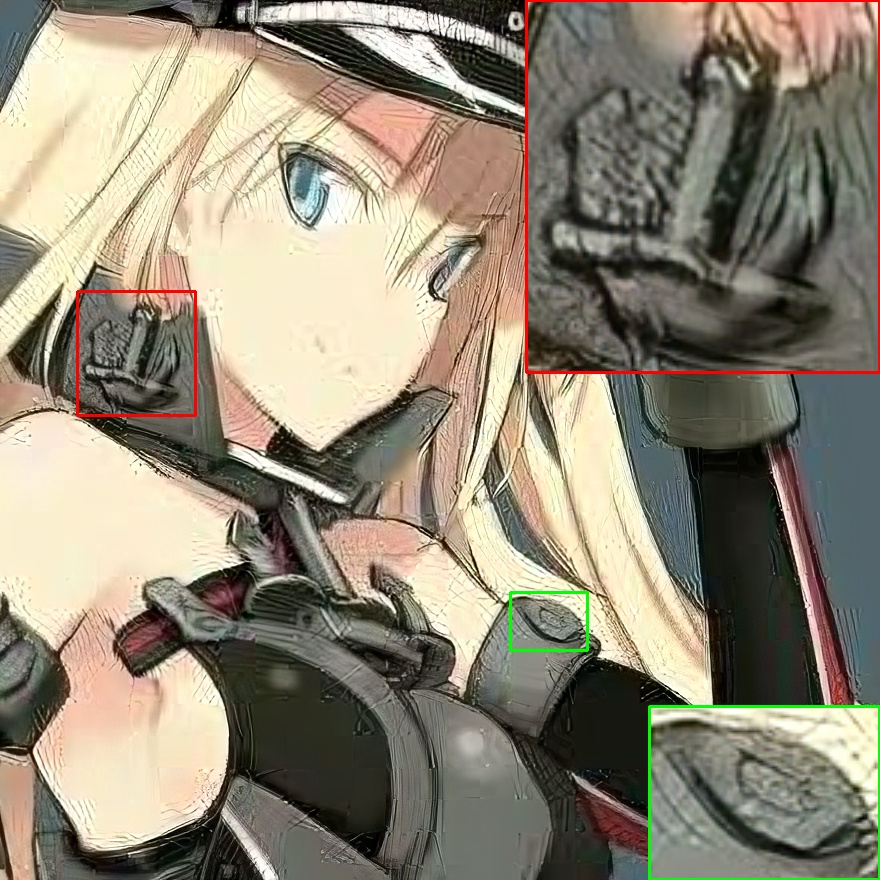}\\ESRGAN~\cite{esrgan}
\end{minipage}
\begin{minipage}[t]{0.16\linewidth}
\centering
\includegraphics[width=1\columnwidth]{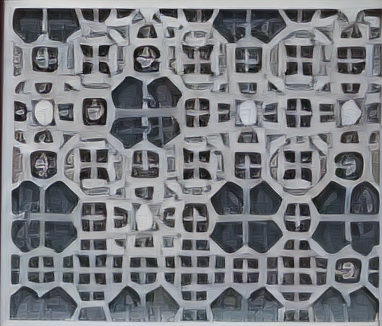}\\
\includegraphics[width=1\columnwidth]{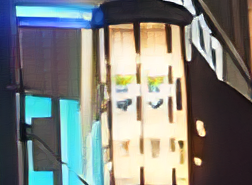}\\
\includegraphics[width=1\columnwidth]{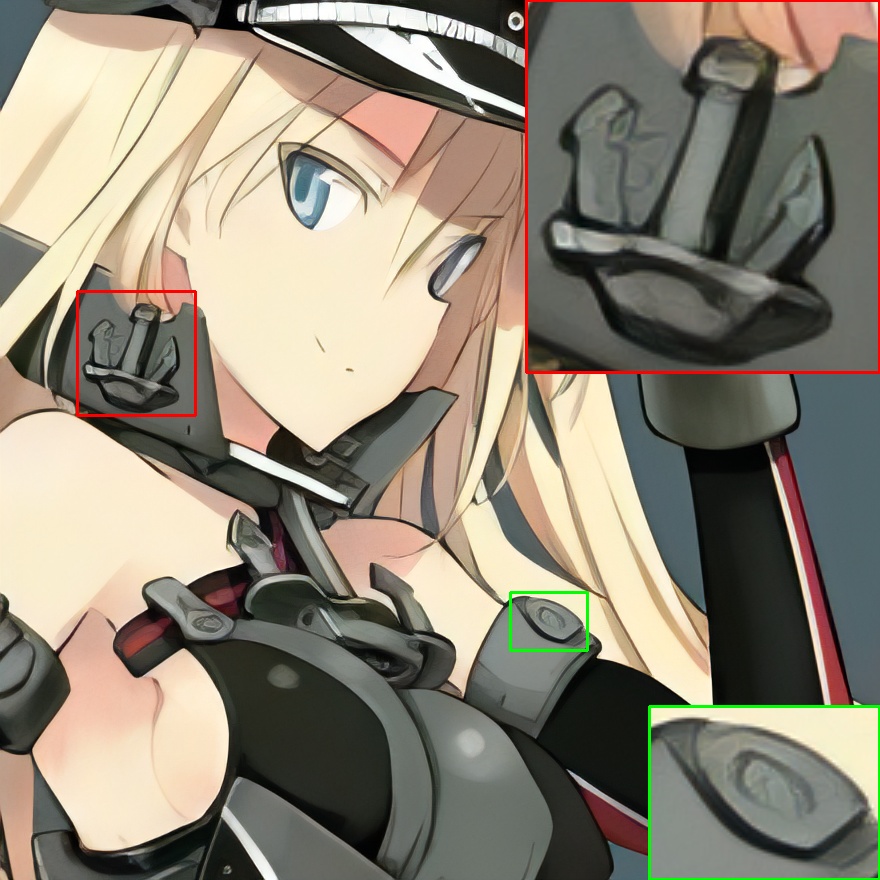}\\BSRGAN~\cite{bsrgan}
\end{minipage}
\begin{minipage}[t]{0.16\linewidth}
\centering
\includegraphics[width=1\columnwidth]{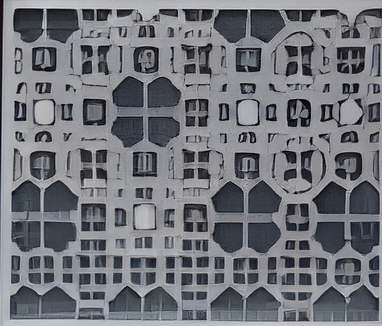}\\
\includegraphics[width=1\columnwidth]{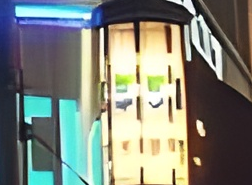}\\
\includegraphics[width=1\columnwidth]{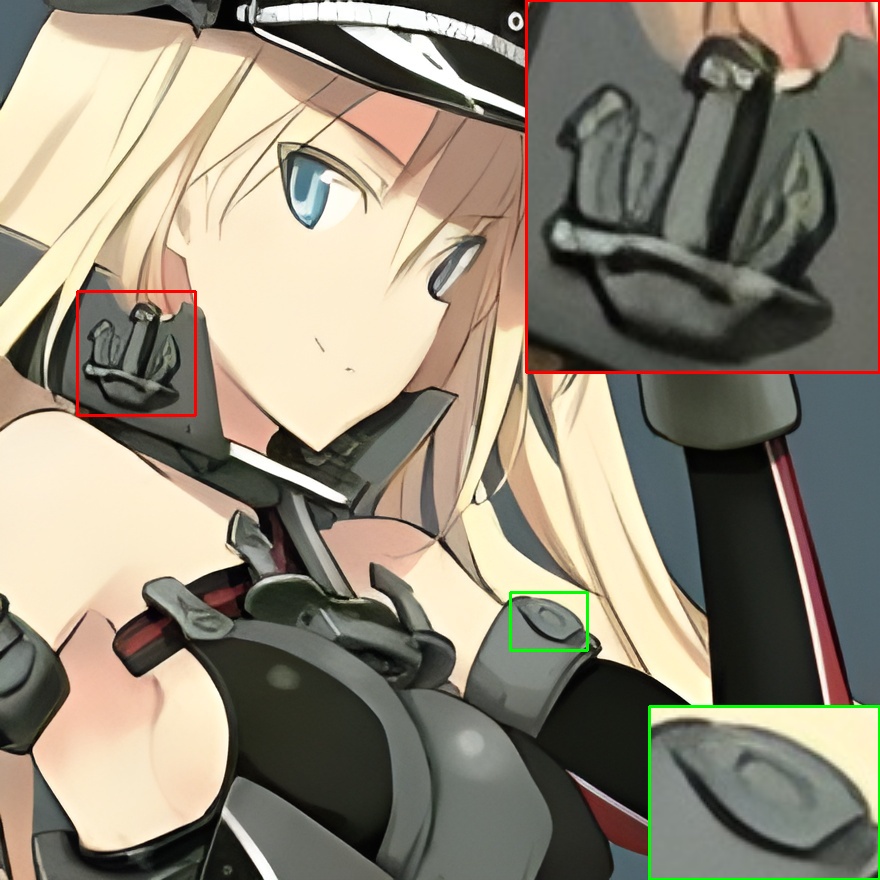}\\ {\tiny RealESRGAN~\cite{realesrgan}}
\end{minipage}
\begin{minipage}[t]{0.16\linewidth}
\centering
\includegraphics[width=1\columnwidth]{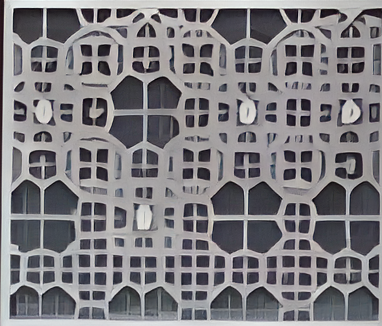}\\
\includegraphics[width=1\columnwidth]{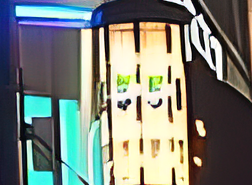}\\
\includegraphics[width=1\columnwidth]{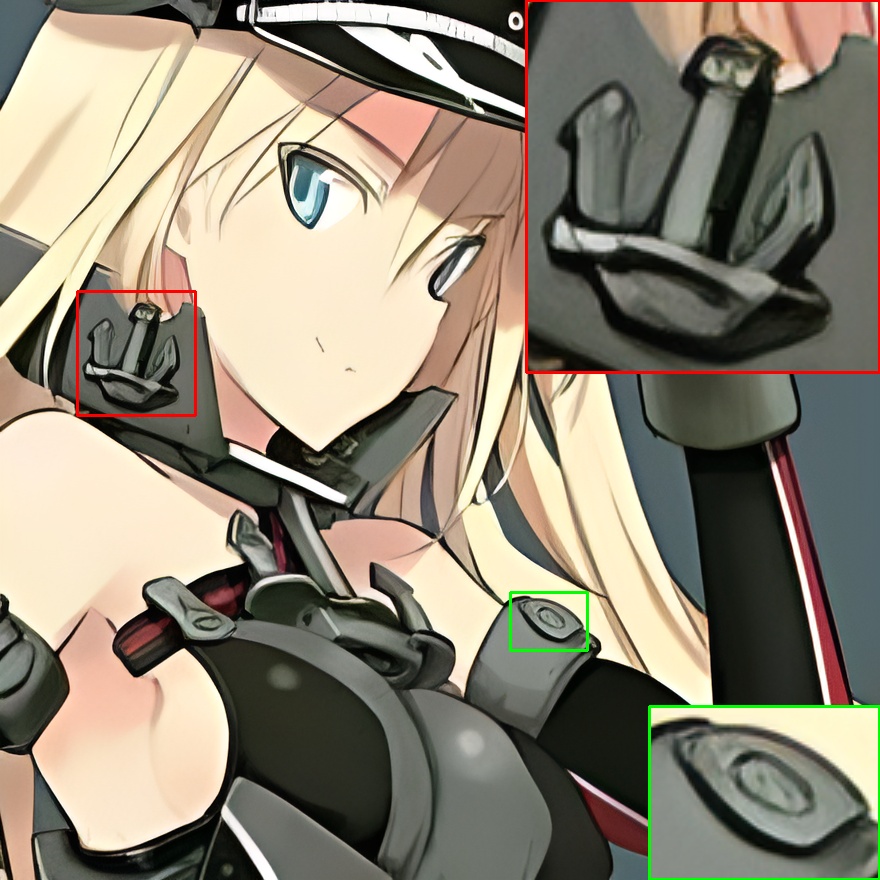}\\MM-RealSR
\end{minipage}
\centering
\caption{Visual comparison between our proposed MM-RealSR and several recent methods on real-world super-resolution. The degradation scores are estimated by our UDEM without specific adjustment during inference.}
\label{v_cp} 
\end{figure*}

\begin{table}[t]
\caption{Quantitative comparisons (NIQE/LPIPS/DISTS) on RealSRSet~\cite{realsr} and AIM19~\cite{aim19} test sets. Best results are \textbf{highlighted}. We compare our MM-RealSR with both modulation (Mod.) and non-modulation (Non-mod.) methods. Real-ESRGAN~\cite{realesrgan} is the upper-bound model of our MM-RealSR.} 
\scriptsize
\renewcommand\tabcolsep{0.6pt}
\centering
\begin{tabular}{c c c c c c c c}
\hline
& & RealSR~\cite{smreal2} & ESRGAN~\cite{esrgan} & CUGAN~\cite{cresmd2} & BSRGAN~\cite{bsrgan} & Upper bound~\cite{realesrgan} & MM-RealSR\\
\hline
\multicolumn{2}{c}{Setting} & Non-mod. & Non-mod. & Mod. & Non-mod. & Non-mod. & For both\\
\hline
\hline
\multirow{3}{*}{\makecell[c]{Real\\SRSet}} & NIQE $\downarrow$ & 8.31 & 8.23 & 6.17 & 4.62 & \textbf{4.53} & 4.62\\
& LPIPS $\downarrow$ & 0.4047 & 0.4054 & 0.3757 & 0.3647 & 0.3640 & \textbf{0.3606}\\
& DISTS $\downarrow$ & 0.2201 & 0.2190 & 0.1764 & \textbf{0.1535} & 0.1545 & 0.1664\\
\hline
\multirow{3}{*}{AIM19} & NIQE $\downarrow$ & \textbf{2.86} & 3.60 & 4.77 & 3.92 & 3.60 & 3.61\\
& LPIPS $\downarrow$ & 0.5566 & 0.5558 & 0.4635 & 0.4048 & 0.3957 & \textbf{0.3948}\\
& DISTS $\downarrow$ & 0.2325 & 0.2336 & 0.1944 & 0.1596 & 0.1545 & \textbf{0.1507}\\
\hline
\end{tabular}
\label{qc}
\end{table}

\subsection{Evaluation for Restoration Performance}
\textbf{Comparison details.} In this part, we fully compare our proposed MM-RealSR with recent modulation (\textit{i.e.}, CUGAN~\cite{cresmd2}) and non-modulation methods. The non-modulation methods include several well-known SR methods (\textit{e.g.}, RealSR \cite{smreal2} and ESRGAN \cite{esrgan}) and some recent top-performing methods (\textit{e.g.}, BSRGAN \cite{bsrgan} and Real-ESRGAN \cite{realesrgan}). Real-ESRGAN is also the theoretical upper bound of our proposed MM-RealSR, as MM-RealSR needs to learn another objective for interactive modulation. Note that CUGAN must be provided with the degradation levels ($S_b,\ S_n \in [0,\ 1]$), which are not available in real-world test sets, \textit{i.e.}, RealSRSet and AIM19. Thus, we divide the noise and blur degradation levels into $11$ points uniformly distributed between 0 and 1, and then traversed all cases ($121$ cases in total). We select the best result (lowest LPIPS) for each image. Unlike the above two settings, our MM-RealSR can be used either with the degradation scores estimated by our UDEM module or with the user input scores. In this part, to make a fair comparison, the input degradation score of our MM-RealSR is provided by UDEM without manual adjustment. 

\textbf{Comparison results.} The quantitative comparison is presented in Tab.~\ref{qc}. One can see that compared with other methods, our proposed MM-RealSR achieves excellent performance even with interactive modulation ability. 
Especially, our performance is comparable or even better than the upper-bound model Real-ESRGAN. Even selecting the best result for each image, the performance of CUGAN is still inferior to our proposed MM-RealSR. It demonstrates the incapability of existing modulation methods in real-world scenarios. Though RealSR and ESRGAN achieve better performance on NIQE on the AIM19 test set, the reconstruction quality of these two methods is unsatisfactory, as shown in Fig.~\ref{v_cp}. 
From the visual comparison in Fig.~\ref{v_cp}, we can find that our MM-RealSR can achieve better restoration quality with vivid details, demonstrating that our approach can estimate the degradation scores to guide the restoration process in real-world scenarios.


\begin{figure*}[t]
\centering
\small 
\begin{minipage}[t]{.9\linewidth}
\centering
\includegraphics[width=1\columnwidth]{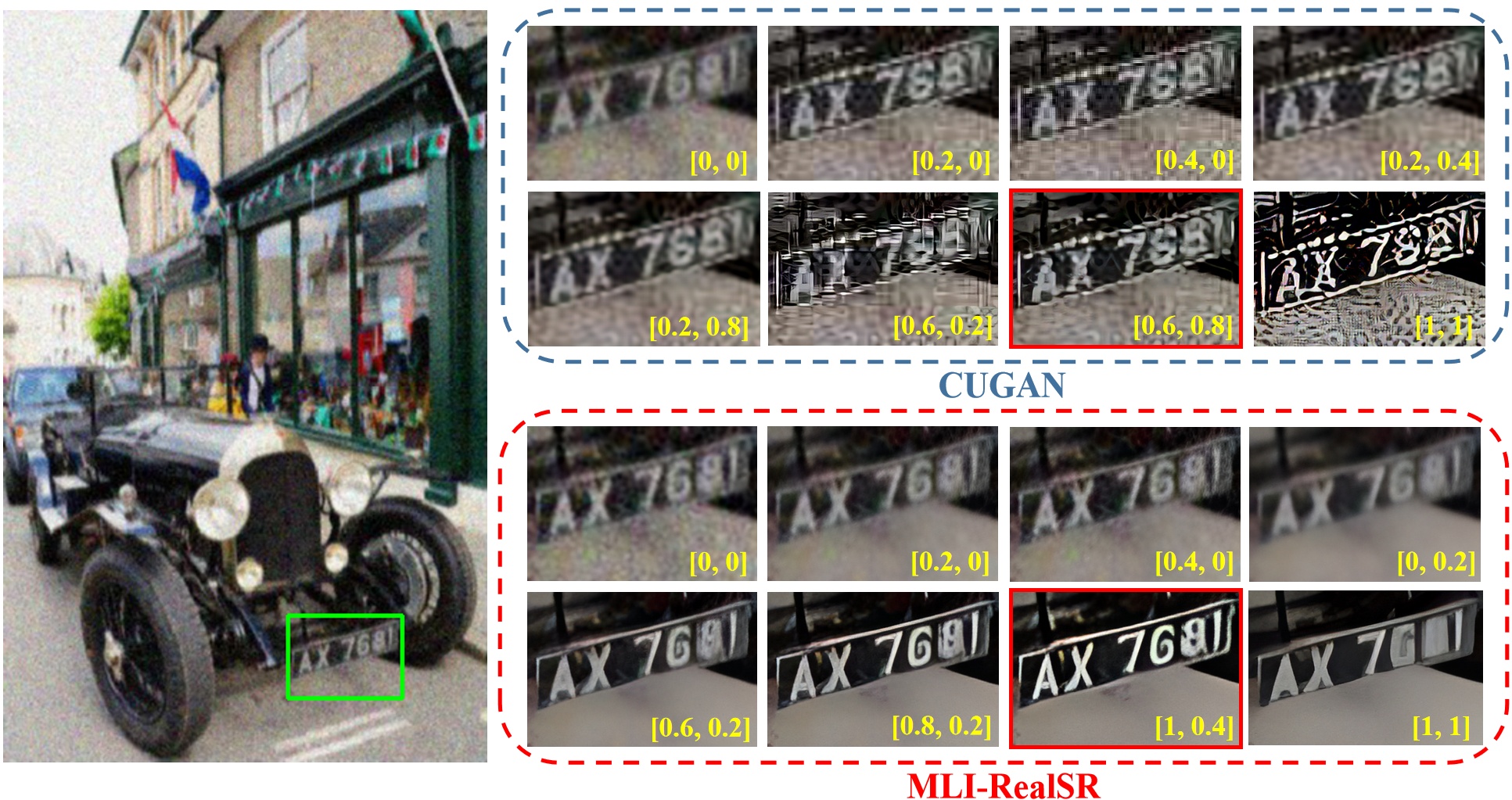}
\end{minipage}
\centering
\caption{The evaluation of the modulation performance on real-world super-resolution ($\times 4$). The degradation scores (\textit{i.e.}, $[S_b,\ S_n]$) are manually selected
for CUGAN~\cite{cresmd2} and our proposed MM-RealSR to present the visual quality and adjusting ability. More comparison videos are in the appendix.
}
\label{v_rwm} 
\end{figure*}

\subsection{Real-World Modulation Performance}
\textbf{Comparison with CUGAN.}
In this part, 
We compare our method with the most recent modulation-based SR method CUGAN~\cite{cresmd2} on real-world data. We visualize the modulation process of our MM-RealSR and CUGAN in Fig.~\ref{v_rwm}. 
The best visual result of each method is labeled with a red box
We can find that the CUGAN trained with simple degradations can not deal with real-world scenarios. On the contrary, our MM-RealSR has better reconstruction quality with a larger modulation range in real-world applications. 


\begin{table}[t]
\caption{The quantitative comparison between our MM-RealSR and the segmented modulation approach. The quantitative results are evaluated on RealSRSet test set~\cite{realsr}.
}
\footnotesize
\renewcommand\tabcolsep{3pt}
\centering
\begin{tabular}{c c c c c c c}
\hline
 & Segment-1 & Segment-2 & Segment-3 & Segment-4 & Best-LPIPS & MM-RealSR\\
\hline
\hline
LPIPS $\downarrow$ & 0.3957 & 0.3910 & 0.3950 & 0.4125 & 0.3905 & \textbf{0.3662}\\
DISTS $\downarrow$ & 0.1764 & 0.1820 & 0.1910 & 0.1942 & 0.1760 & \textbf{0.1632}\\
\hline
\end{tabular}
\label{naive}
\end{table}

\begin{figure*}[t]
\centering
\footnotesize
\begin{minipage}[t]{0.16\linewidth}
\centering
\includegraphics[width=1\columnwidth]{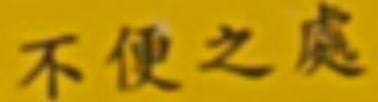}\\ 
\includegraphics[width=1\columnwidth]{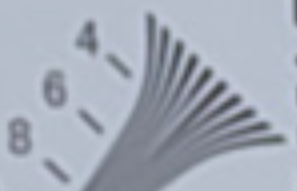}\\Bicubic $\times 4$
\end{minipage}
\begin{minipage}[t]{0.16\linewidth}
\centering
\includegraphics[width=1\columnwidth]{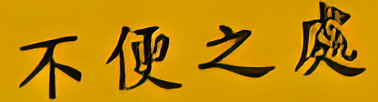}\\ 
\includegraphics[width=1\columnwidth]{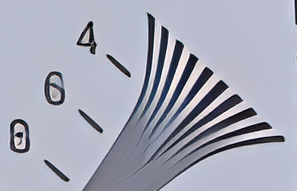}\\Segment-1
\end{minipage}
\begin{minipage}[t]{0.16\linewidth}
\centering
\includegraphics[width=1\columnwidth]{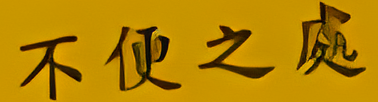}\\ 
\includegraphics[width=1\columnwidth]{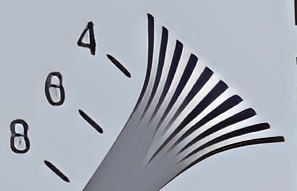}\\Segment-2
\end{minipage}
\begin{minipage}[t]{0.16\linewidth}
\centering
\includegraphics[width=1\columnwidth]{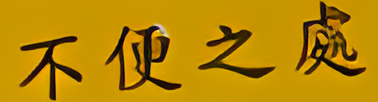}\\ 
\includegraphics[width=1\columnwidth]{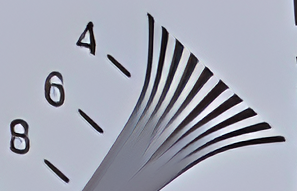}\\Segment-3
\end{minipage}
\begin{minipage}[t]{0.16\linewidth}
\centering
\includegraphics[width=1\columnwidth]{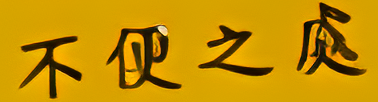}\\ 
\includegraphics[width=1\columnwidth]{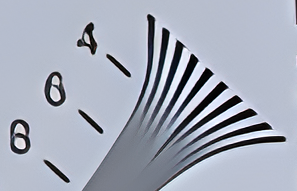}\\Segment-4
\end{minipage}
\begin{minipage}[t]{0.16\linewidth}
\centering
\includegraphics[width=1\columnwidth]{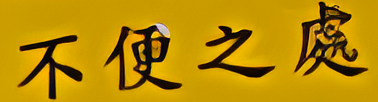}\\ 
\includegraphics[width=1\columnwidth]{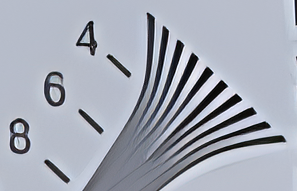}\\MM-RealSR
\end{minipage}
\centering
\caption{Visual comparison between our MM-RealSR and the segmented modulation approach. The input degradation scores of our MM-RealSR are estimated by UDEM.
}
\label{v_naive} 
\end{figure*}

\textbf{Further Discussions.}
A direct interactive modulation method for RWSR is to train several models with different degradation levels. The modulation can then be achieved by performing inference on those models and selecting the most satisfactory result.
To compare our method with this case, we uniformly divide the degradations in training Real-ESRGAN~\cite{realesrgan} into four segments, corresponding to four models.
In the evaluation, we select the best result (lowest LPIPS) among the four results for each sample. We experiment on the RealSRSet~\cite{realsr} and show the results in Tab.~\ref{naive} and Fig.~\ref{v_naive}. Obviously, the performance of such a simple approach is inferior to our method both quantitatively and qualitatively. Besides, this simple method only has four points for modulation.


\section{Conclusion}
In this paper, we present the first attempt to study the interactive modulation in real-world super-resolution through a novel metric learning based strategy. Specifically, we utilize metric learning to map the unquantifiable degradation level in real-world scenarios to a metric space and train it in an unsupervised manner. Equipped with the metric space, we can get a relative ranking score of the unquantifiable real-world degradation to guide the restoration and modulation processes. We also propose an anchor point strategy to constrain the distribution in the metric space. To adapt to the complex degradations in real-world scenarios, we also carefully design controllable dimensions in which each degradation factor has a more general degradation meaning. Extensive experiments demonstrate that our proposed MM-RealSR achieves excellent modulation and restoration performance in real-world super-resolution.

\textbf{Limitations.} Our MM-RealSR adopts the complex and high-order degradation synthesis~\cite{bsrgan,realesrgan} to simulate the real-world degradations.
However, it is still hard to directly apply our method to \text{train} on real samples.
There are also some failure cases (\textit{e.g.}, local jitter in Fig.~\ref{dist}, and our UDEM regards some art effects such as bokeh as degradation).
Our work makes the first attempt towards interactive modulation in RWSR, and there is still a long way with challenges.

\textbf{Acknowledgement.} This work was partially supported by the Shenzhen Fundamental Research Program (No.GXWD20201231165807007-20200807164903001),  National Natural Science Foundation of China (61906184, U1913210), and the Shanghai Committee of Science and Technology, China (Grant No. 21DZ1100100).

\bibliographystyle{splncs04}
\bibliography{egbib}

\end{document}